\documentclass[sigconf]{acmart}

\copyrightyear{2021}
\acmYear{2021}
\setcopyright{rightsretained}
\acmConference[KDD '21] {Proceedings of the 27th ACM SIGKDD Conference on Knowledge Discovery and Data Mining}{August 14--18, 2021}{Virtual Event, Singapore.}
\acmBooktitle{Proceedings of the 27th ACM SIGKDD Conference on Knowledge Discovery and Data Mining (KDD '21), August 14--18, 2021, Virtual Event, Singapore}
\acmPrice{}
\acmISBN{978-1-4503-8332-5/21/08}
\acmDOI{10.1145/3447548.3467326}

\AtBeginDocument{%
  \providecommand\BibTeX{{%
    \normalfont B\kern-0.5em{\scshape i\kern-0.25em b}\kern-0.8em\TeX}}}

\usepackage[libertine]{newtxmath}  

\usepackage{amsmath,amsfonts,bm,mathtools}

\def\eqref#1{equation~\ref{#1}}

\def\1{\bm{1}}

\DeclareMathAlphabet{\mathsfit}{\encodingdefault}{\sfdefault}{m}{sl}
\SetMathAlphabet{\mathsfit}{bold}{\encodingdefault}{\sfdefault}{bx}{n}

\newtheorem{dfn}{Definition}

\theoremstyle{remark}

\def\L{\mathcal{L}}
\def\F{\mathcal{F}}
\def\Lhat{\hat{\mathcal{L}}}
\def\Fhat{\hat{\mathcal{F}}}

\def\Y{\mathcal{Y}}

\usepackage{graphicx}
\usepackage[font=default,textfont=default,labelfont=bf]{caption}
\usepackage{subcaption}
\usepackage{multirow}
\usepackage[linesnumbered, ruled,vlined]{algorithm2e}
\SetCommentSty{mycommfont}
\SetKwComment{Comment}{$\triangleright$\ }{}
\let\oldnl\nl
\newcommand{\nonl}{\renewcommand{\nl}{\let\nl\oldnl}}

\usepackage{bbm}
\usepackage{amsmath}
\usepackage{enumitem}

\begin{document}
\fancyhead{}  

\title{Understanding and Improving Fairness-Accuracy Trade-offs in Multi-Task Learning}


\author{Yuyan Wang, Xuezhi Wang, Alex Beutel, Flavien Prost, Jilin Chen, Ed H. Chi}
\affiliation{%
  \institution{Google Research, USA}
\{yuyanw,xuezhiw,alexbeutel,fprost,jilinc,edchi\}@google.com}








\begin{abstract}
As multi-task models gain popularity in a wider range of machine learning applications, it is becoming increasingly important for practitioners to understand the fairness implications associated with those models. 
Most existing fairness literature focuses on learning a single task more fairly, while how ML fairness interacts with multiple tasks in the joint learning setting is largely under-explored. In this paper, we are concerned with how group fairness (e.g., equal opportunity, equalized odds) as an ML fairness concept plays out in the multi-task scenario. In multi-task learning, several tasks are learned jointly to exploit task correlations for a more efficient inductive transfer. This presents a multi-dimensional Pareto frontier on (1) the trade-off between group fairness and accuracy with respect to each task, as well as (2) the trade-offs across multiple tasks. We aim to provide a deeper understanding on how group fairness interacts with accuracy in multi-task learning, and we show that traditional approaches that mainly focus on optimizing the Pareto frontier of multi-task accuracy might not perform well on fairness goals. We propose a new set of metrics to better capture the multi-dimensional Pareto frontier of fairness-accuracy trade-offs uniquely presented in a multi-task learning setting. 
We further propose a Multi-Task-Aware Fairness (MTA-F) approach to improve fairness in multi-task learning. Experiments on several real-world datasets demonstrate the effectiveness of our proposed approach.
\end{abstract}





\maketitle

\section{Introduction}
\label{s1_introduction}

Over the past years, multi-task deep learning has gained popularity through its success in a wide range of applications, including natural language processing \cite{collobert2008unified}, computer vision \cite{girshick2015fast, ren2015faster}, and online recommendation systems \cite{bansal2016ask, ma2018modeling}. 
Despite the popularity of using multi-task learning in various real-world applications, the understanding of fairness under such a framework is largely under-explored. Most existing work on fairness focuses on a single-task learning setting, including work on fair representation learning \cite{zemel13, pmlr-v108-tan20a, zhao2020, Madras2018LearningAF, beutel17}, and fairness mitigation \cite{gupta16, gupta19,DBLP:conf/icml/AgarwalBD0W18, pmlr-v54-zafar17a, beutel2019putting}.

Here we focus on the widely used notion of group fairness, \emph{equal opportunity} and \emph{equalized odds} proposed by \cite{hardt2016equality}, which aims at closing the gap of true positive rates and false positive rates across different groups. In a single-task learning setting, the Pareto frontier can be characterized by trading-off some notions of accuracy and group fairness over a single task \cite{Menon2018TheCO, liobait2015OnTR}.

Fairness in multi-task learning poses new challenges and the need of characterizing a multi-dimensional Pareto frontier. In a traditional multi-task learning setting where fairness is not taken into consideration, people focus on optimizing the Pareto frontier of multiple accuracies across tasks. Instead, our work aims at demystifying the multi-dimensional trade-off and improving fairness on top of accuracy objectives for multi-task learning problems.

First, we show analysis that traditional multi-task learning objectives might not correlate well with fairness goals, thus demonstrating that fairness needs to be better handled in multi-task learning settings.
To better evaluate fairness in multi-task learning, we then propose a new set of metrics to capture the multi-dimensional nature of the Pareto frontier.
This includes two sets of trade-offs, one that captures the fairness vs. accuracy trade-off with respect to each task, and the other captures the trade-offs across multiple tasks.
We show that due to the intrinsic differences in various tasks, a simple aggregation of the accuracy or fairness metrics of each task might not be a fair assessment of the entire system. Specifically, when different tasks suffer from fairness issues to different extents (and thus have fairness metrics in different scales), we propose a set of normalized metrics (Average Relative Fairness Gap and Average Relative Error, Section \ref{sec:measuring}) that better captures the overall trade-off between fairness and accuracy.

Finally, we propose a data-dependent mitigation method, \emph{Multi-Task-Aware Fairness treatment (MTA-F)}, that improves the Pareto frontiers for fairer multi-task learning. The main idea is to decompose the remediation treatments for different components of the multi-task model architecture according to label distributions across all tasks. By doing so, MAT-F allows flexible mitigation over (1) the representation learning shared by all tasks, and (2) the non-shared sub-networks specific to each task. This effectively enables more efficient use of limited model capacity across multiple tasks over both accuracy and fairness goals. Compared with a baseline method which is not data-dependent and the same fairness loss is applied to all parts of the multi-task models, MTA-F improves the Pareto efficiency when trading-off fairness vs. accuracy across multiple tasks.
Experiment results on multiple real-world datasets demonstrate the effectiveness of MTA-F in multi-task applications.

To summarize, our contributions are:
\begin{itemize}[noitemsep,topsep=0pt,nosep]
    \item \textbf{Problem Framing:} We provide insights on how fairness interacts with accuracy in multi-task learning. Notably, traditional approaches that focus on optimizing the Pareto frontier of multi-task accuracy may not correlate well with equal opportunity goals.
    \item \textbf{New Metrics:} We propose a new set of metrics that captures the multi-dimensional fairness-accuracy trade-offs uniquely presented in a multi-task learning setting. 
    \item \textbf{New Mitigation:} We provide a data-dependent multi-task fairness mitigation approach, MTA-F, which decomposes fairness losses for different model components by exploiting task relatedness and the shared architecture for multi-task models. 
\end{itemize}

\section{Related Work}
\label{s2_relatedwork}
\textbf{Fairness metrics}. 
A number of fairness metrics have been proposed in existing works, with the focus mostly on fairness in classification. \emph{Demographic parity} \cite{hardt2016equality, liobait2015OnTR} requires that all subgroups receive the same proportion of positive outcomes. Although being adopted as the definition of fairness in a series of works \cite{calders2009building, johndrow2019algorithm, kamiran2009classifying}, demographic parity may be unrealistic to achieve in practice, especially in the common scenarios where the base rates differ across subgroups \cite{zhao2019inherent}. Another set of commonly used fairness metrics, \emph{equal opportunity} and \emph{equalized odds} \cite{hardt2016equality}, has also been widely adopted in measuring discrimination against protected attributes. Instead of requiring equal outcomes, equal opportunity and equalized odds require equal true positive rates and false positive rates across different subgroups, a somewhat more realistic fairness notion in a wide range of applications in practice. For our work, we focus on equal opportunity and equalized odds as our notion of fairness.

\noindent \textbf{Fairer representation learning}. 
One way to address fairness in machine learning is through fairer representation learning \cite{zemel13, pmlr-v108-tan20a, zhao2020}, with the goal being obfuscating information about the protected group membership in the learned representations.
Fairer representation learning can be achieved adversarially \cite{Madras2018LearningAF, beutel17}. Moreover, \citet{Schumann2019TransferOM} propose transfer learning in the representation space to adapt fairness to a new domain.

\noindent \textbf{Fairness mitigation}.
There is a large body of work on mitigating fairness issues in a single-task learning setting. Examples include pre-processing the data embeddings to help downstream models be trained more fairly \cite{tolga2016}, and post-processing model's predictions to improve fairness \cite{calibration17,post19}.
In addition, intervening the model training process has also been popular, including adding fairness constraints  \cite{gupta16, gupta19,DBLP:conf/icml/AgarwalBD0W18} or regularization \cite{pmlr-v54-zafar17a,JMLR:v20:18-262,beutel2019putting}.
Different from existing works that add fairness constraints directly into the model's objective for single-task learning, our work exploits the multi-task model architecture with a decomposition and redistribution of the fairness constraints for fairer multi-task learning.

\noindent \textbf{Multi-task learning}. 
A common approach for multi-task learning is to design a parameterized model that shares a subset of parameters across different tasks \cite{ruder2017overview}. The benefits of such shared architectures are numerous. 
First, it exploits task relatedness with inductive bias learning \citep{caruana1997multitask, baxter2000model}. Learning a shared representation across related tasks is beneficial especially for harder tasks or tasks with limited training examples.
Second, by forcing tasks to share model capacity, it introduces a regularization effect and improves generalization. 
Third, it offers a compact and efficient form of modeling which enables training and serving multiple prediction quantities for large-scale systems. Learning multiple tasks together can improve the performance of some tasks but hurt others \cite{standley2019tasks}. Existing literature in multi-task learning focus on reducing the task training conflicts \cite{kendall2018multi, ma2018modeling, yu2020gradient, chen2020just, killamsetty2020reweighted} and improving the Pareto frontier. 

\noindent \textbf{Fairness in multi-task learning}.
Fairness has been mostly studied in single-task settings, and little work has been done in the context of multi-task learning. However, as multi-task learning becomes prevalent in state-of-the-art model designs \cite{Zhang2017ASO, Ruder2017AnOO}, it is important for practitioners to understand how fairness interacts with multi-task learning. 
\citet{d2020underspecification} provide insights on how pure multi-task learning can have unintended effects on fairness.
\citet{mtl_regression} study fairness in multi-task regression models and uses a rank-based non-parametric independence test to improve fairness in ranking.
\citet{mtl_fair} propose to use multi-task learning enhanced with fairness constraints to jointly learn classifiers that leverage information between sensitive groups.
\citet{zhao2020primal} study fair meta-learning.
\citet{compositional} show that fairness might not compose in multi-component recommenders. 

\noindent \textbf{Fairness-accuracy trade-off and Pareto fairness}.
Finally, a great amount of work has shown that fairness usually comes with a trade-off over accuracy, e.g., \cite{liobait2015OnTR, Menon2018TheCO, Zhao2019InherentTI, DBLP:journals/corr/abs-1911-06935}, again mostly under a single-task setting. In single-task learning, the Pareto frontier can be characterized by trading-off some notions of the accuracy and group fairness over a single task. For example, the objective is generally in the form of $\L(t) + \lambda \F(t)$, where $\L(t)$ represents the loss over accuracy for task $t$, $\F(t)$ is the fairness loss, and $\lambda$ is a parameter trading-off accuracy and fairness.  
\citet{ananth2019} explore the pareto-efficient fairness and show that it achieves Pareto levels in accuracy for all subgroups for a single-task.
\citet{martinez2020minimax} and \citet{diana2020convergent} propose to find Pareto optimal solutions for minimax group fairness in which fairness is measured by worst-case single-task accuracies across groups.

In a traditional multi-task learning setting where fairness is not taken into consideration, people focus on optimizing the Pareto frontier of multiple accuracies across tasks, where the objective can be usually written as $\alpha \L(t_1) + (1-\alpha) \L(t_2)$ assuming two tasks $t_1, t_2$ and $\alpha \in [0,1]$ is the weight on task $t_1$. Fairness in multi-task learning poses the need of characterizing a multi-dimensional Pareto frontier: For each single-task, there is a trade-off between fairness and accuracy, while at the same time this trade-off needs to be further balanced across all the tasks.
For example, in the case with two tasks, one can extend the fairness losses from single-task learning to multi-task learning, by formulating the objective as: 
\[
\alpha[\L(t_1) + \lambda_1 \F(t_1)] + (1-\alpha) [\L(t_2) + \lambda_2 \F(t_2)].
\]
The number of objectives and metrics of interest is doubled compared with the case with single-task learning or multi-task learning without fairness treatments, hence introducing challenges in understanding and optimizing the multi-dimensional Pareto frontier. 

In this work, we focus on studying this fairness-accuracy trade-off under the \emph{multi-task} learning setting which has been largely under-explored before.
In particular, we aim to provide a better understanding on the following key questions: 1) How does inductive transfer \cite{baxter2000model} in multi-task learning implicitly impacts fairness? 2) How do we efficiently measure the fairness-accuracy trade-off for multi-task learning, given the multi-dimensional nature of the trade-off surface? 3) Are we able to achieve a better Pareto efficiency in fairness across multiple tasks, by exploiting task relatedness and shared architecture that's specific to multi-task learning?

\section{Problem Definition and Examples}
\label{sec:problem}

\subsection{Multi-Task Learning}
\label{section3.1}
\begin{figure}[hb]
    \centering
    \includegraphics[width=0.45\linewidth]{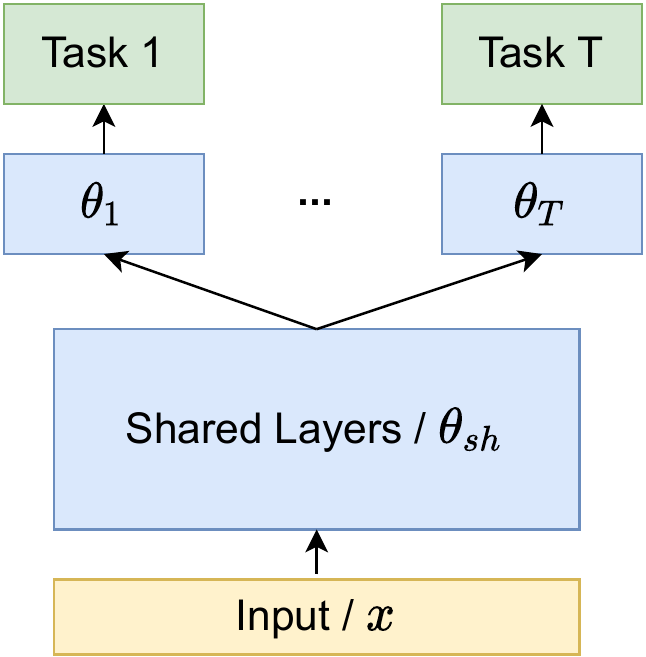}
    \caption{Shared-bottom architecture for a multi-task model.}
    \label{fig:shared-bottom-mtl}
\end{figure}

Suppose there are $T$ tasks sharing an input space $\mathcal{X}$. Each task has its own task space ${\{\Y^t\}}_{i=1}^T$ . A dataset of $n$ i.i.d. examples from the input and task spaces is given by ${\{(x_i, y^1_i, ... ,y^T_i)\}}_{i=1}^n$, where $y^t_i$ is the label of the $t$-th task for example $i$. We assume a multi-task model parameterized by $\theta\in\Theta$. Here $\theta = (\theta_{sh}, \theta_1,...,\theta_T)$ includes shared-parameters $\theta_{sh}$ and task-specific parameters $\theta_1,...,\theta_T$. Let $f_t(\cdot, \cdot): \mathcal{X}\times\Theta \rightarrow \mathcal{Y}^t$ be the model function and $\L_t(\cdot, \cdot): \mathcal{Y}^t\times\mathcal{Y}^t \rightarrow \mathbb{R}^{+}$ be the loss function for the $t$-th task. This generalizes easily to the more general multi-task learning setting where different tasks have different inputs. Figure \ref{fig:shared-bottom-mtl} is an illustration of the typical shared-bottom architecture for a multi-task model, as used in a wide range of applications \cite{caruana1997multitask, Ruder2017AnOO, collobert2008unified, ma2018modeling}.

Let $\Lhat_t(\theta) \coloneqq \frac{1}{n}\sum_{i=1}^n \L_t(f_t(x_i;\theta_{sh}, \theta_t), y^t_i)$ be the empirical loss for the $t$-th task, where we drop the dependency on $x$ and $y$ for ease of notation. The optimization for multi-task learning can then be formulated as a joint optimization of a vector-valued loss function:
\begin{equation}
\label{eqn:3.1.1}
\min_{\theta} (\Lhat_1(\theta), ..., \Lhat_T(\theta))^\top.
\end{equation}
It is unlikely that a single $\theta$ optimizes all objectives simultaneously. The solution to (\ref{eqn:3.1.1}) is therefore a set of points which represent different trade-off preferences. Formally, solution $\theta_a$ is said to \emph{Pareto dominate} solution $\theta_b$ if $\Lhat_t(\theta_a) \leq \Lhat_t(\theta_b) ,\forall t $ and there exist at least one task $j$ such that the inequality is strict. A solution $\theta$ is called \textit{Pareto optimal} or \textit{Pareto efficient} if there is no solution $\theta' \neq \theta$ such that $\theta'$ dominates $\theta$. The \emph{Pareto frontier} is the set of all Pareto optimal solutions.

The most popular choice for the minimization objective is a scalarization of the empirical loss vector \cite{marler2004survey}:
\begin{equation}
\label{eqn:3.1.2}
\Lhat(\theta)\coloneqq\sum_{t=1}^T w_t\Lhat_t(\theta),
\end{equation}
where $\{w_t\}_{t\in\{1,...,T\}} \geq 0$ are weights for individual tasks, which controls the trade-off among different tasks. 

\subsection{Fairness Definition and Metrics}
\label{section3.2}
To measure fairness, we assume each example belongs to a subgroup defined by the sensitive attribute $A$ (e.g. gender). The sensitive attribute information could be available only for a fraction of the training data. For ease of notation we assume the sensitive attribute is a binary variable (e.g. $A=1$ for female, $A=0$ for male).      

We focus on two widely used notions of group fairness, equal opportunity and equalized odds \cite{hardt2016equality}. Equal opportunity is defined as equality of true positive rates (TPR) or false positive rates (FPR) across groups\footnote{The original paper \cite{hardt2016equality} frames equal opportunity as equality of TPR, but equality of FPR is also widely used in many literature and applications.}. Equalized odds is defined as the predicted outcome $\hat{Y}$ being independent of the group membership conditional on the true outcome. In the case of classification, it entails equal true positive and false positive rates across different groups. 
The discrepancies in FPR and TPR are defined as FPR gap and TPR gap:
\begin{align} 
\label{eqn3.2.1}
\begin{split}
FPRGap &= |P(\hat{Y}=1|Y=0, A=0) - P(\hat{Y}=1|Y=0, A=1)|, \\
TPRGap &= |P(\hat{Y}=1|Y=1, A=0) - P(\hat{Y}=1|Y=1, A=1)|.
\end{split}
\end{align} 

We use FPR Gap to measure the deviation from equal opportunity, and FPR gap and TPR gap for equalized odds.

\subsection{Measuring Fairness in Multi-Task Learning}
\label{section3.3}
In this section, we look into how training multiple tasks together can have implications on the fairness of individual tasks.

We look at two datasets with known fairness concerns. The CelebA dataset \cite{liu2015deep} contains public domain images of public figures. It has more than 200K images and each has 40 binary attribute annotations. We pick 2 attributes, \emph{Attractive} and \emph{Smiling} which are known to be biased in gender \cite{denton2019detecting}, as the tasks for multi-task learning. 
The UCI-Adult dataset\footnote{https://archive.ics.uci.edu/ml/datasets/adult} \cite{Dua:2019} contains census information of over 48,000 adults from the 1994 Census, where we define  $Income >\$50,000$ or $\leq\$50,000$ as Task 1, and $Capital\; Gain > 0$ or not as Task 2. Details of the model can be found in Appendix \ref{sup:experiment_setup}. For both datasets, we focus on gender as the sensitive attribute.


\begin{table}[b]
 \begin{subtable}{0.45\textwidth}
  \begin{center}
    \begin{tabular}{c|cccc}
      \hline\hline
                  & T1 Error & T1 FPR Gap  & T2 Error & T2 FPR Gap \\
      \hline
     STL-T1     & 0.2030  & 0.2716  & -       & -        \\

     STL-T2     & -       & -      & 0.0784  & 0.0145    \\

     MTL        & 0.2035  & 0.2846  & 0.0783  & 0.0137    \\
     \hline
     Difference & +0.24\%  & \textbf{+4.78\%} & -0.08\% & \textbf{-5.39\%} \\

      \hline\hline
    \end{tabular}
    \caption{\textbf{CelebA}: MTL hurts Task 1 fairness but improves Task 2 fairness.}
    \label{tab3.3.1a}
  \end{center}
 \end{subtable}
 
 \begin{subtable}{0.45\textwidth}
  \begin{center}
    \begin{tabular}{c|cccc}
      \hline\hline
                  & T1 Error & T1 FPR Gap  & T2 Error & T2 FPR Gap \\
      \hline
     STL-T1     & 0.1659  & 0.1200   & -      & -    \\

     STL-T2     & -       & -       & 0.1313  & 0.0661    \\

     MTL        & 0.1656  & 0.1205  & 0.1299  & 0.0738    \\
     \hline
     Difference & -0.20\%  & +0.34\% & \textbf{-1.10\%} & \textbf{+11.60\%} \\

      \hline\hline
    \end{tabular}
    \caption{\textbf{UCI-Adult}: MTL improves Task 2 accuracy but hurts its fairness.}
    \label{tab3.3.1b}
  \end{center}
 \end{subtable}
 \caption{Fairness and accuracy metrics for single-task learning and multi-task learning. \textbf{STL-T1}: single-task learning for Task 1; \textbf{STL-T2}: single-task learning for Task 2; \textbf{MTL}: multi-task learning; \textbf{Difference}: percentage difference of MTL and STL metrics. Results are averaged over 100 runs. \textbf{Bold} font means the difference is statistically significant at 95\% confidence level.}
 \label{tab3.3.1}
\vspace{-0.1in}
\end{table}
\begin{figure}[hb]
        \centering
        \begin{subfigure}{0.23\textwidth}
                \centering
                \includegraphics[width=\textwidth]{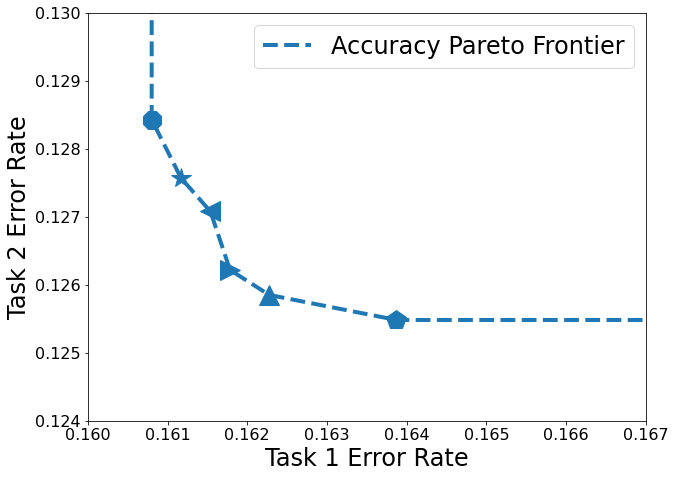}\vspace{-0.1in}
                \caption{Accuracy Pareto frontier.\label{fig:pareto-frontier-uci-a}}
        \end{subfigure}
        \begin{subfigure}{0.23\textwidth}
                \centering
                \includegraphics[width=\textwidth]{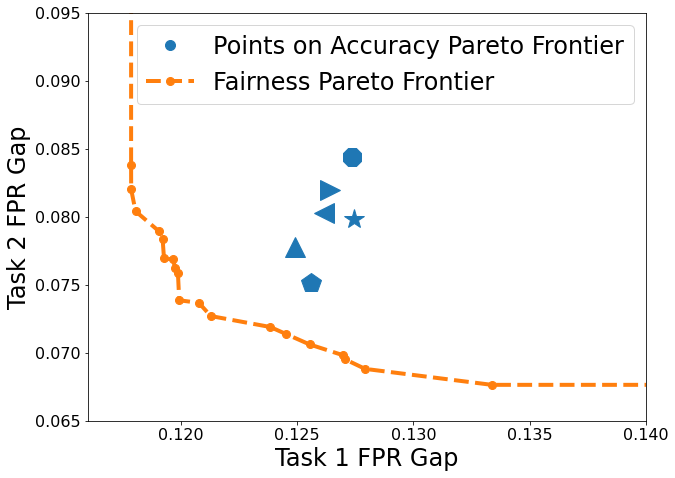}\vspace{-0.1in}
                \caption{Fairness Pareto frontier.\label{fig:pareto-frontier-uci-b}}
        \end{subfigure}
        \vspace{-0.1in}
        \caption{Pareto frontiers on UCI-Adult dataset, where points on lower-left means better accuracy/fairness metrics. Blue points in (b) correspond to the Pareto optimal runs in (a). \label{fig:pareto-frontier-uci-}}
\vspace{-0.1in}
\end{figure}
To understand how training multiple tasks together affects the fairness and accuracy of individual tasks, we compare the FPR gap and error rate on the test data under single-task learning (STL) and multi-task learning (MTL). For STL of each task, we use the same model architecture as in MTL but without the head layers for other tasks in Fig. \ref{fig:shared-bottom-mtl}. Therefore, each task has the same effective model capacity under STL and MTL. 
For MTL, we set $w_1=w_2=0.5$ as the task loss weights in Eq. (\ref{eqn:3.1.2}). 

Table \ref{tab3.3.1} summarizes the numerical results. On the CelebA dataset (Table \ref{tab3.3.1a}), multi-task learning hurts the fairness of Task 1 (\emph{Attractive}) but improves the fairness of Task 2 (\emph{Smiling}). On the UCI-Adult dataset (Table \ref{tab3.3.1b}), while multi-task learning helps the accuracy of Task 2 ($capital\; gain > 0$), it significantly degrades its fairness by increasing its FPR gap by $11.6\%$.  


To further understand this, we perform another 10k runs of the multi-task model on the UCI-Adult dataset, with the task weights $w_1$ and $w_2=1-w_1$ varying in $[0,1]$. Figure \ref{fig:pareto-frontier-uci-} shows the accuracy Pareto frontier and fairness Pareto frontier over those runs. We pick the model runs that are Pareto optimal in accuracies (Fig. \ref{fig:pareto-frontier-uci-a}) and plot their fairness metrics together with the fairness Pareto frontier in Fig. \ref{fig:pareto-frontier-uci-b}. We see that the model runs that are Pareto optimal in accuracies are far from the fairness Pareto frontier, meaning that they are \emph{not} Pareto optimal in terms of fairness. 

The comparisons suggest that training multiple tasks together by simply pooling the accuracy objectives using a shared model architecture may have serious fairness implications---a fact that could be easily overlooked in multi-task applications. Even when only optimizing for accuracy objectives, multi-task learning may have larger impacts on fairness goals than on accuracy goals (Table \ref{tab3.3.1a}), or hurting the fairness of some tasks while benefiting from accuracy gains (Table \ref{tab3.3.1b}). More generally, optimizing for accuracy among multiple tasks may lead to suboptimal trade-offs in fairness (Fig. \ref{fig:pareto-frontier-uci-}). Ignoring these implications in practice may lead to unwanted fairness consequences in multi-task applications.


\section{Metrics for Fairness-Accuracy Pareto Frontier}
\label{sec:measuring}
Here we propose metrics to measure the multi-dimensional trade-off between fairness and accuracy across multiple tasks, and then in the next section propose treatments to improve the trade-off for multi-task models. As shown above, the usual multi-task learning objectives such as accuracy often do not align with fairness goals; A good accuracy trade-off among tasks may have bad fairness implications. Therefore, fairness trade-off must be considered together with accuracy trade-off when evaluating multi-task models. 

To this end, we propose metrics that characterize the Pareto frontier of fairness-accuracy trade-off for multi-task learning. Consider, for example, equal opportunity as the fairness objective and FPR gap as the fairness metric. Let $Err^{(1)}, ..., Err^{(T)}$ be the error rates for the $T$ tasks where $Err = P(\hat{Y} \neq Y)$,
and $FPRGap^{(1)}, ..., FPRGap^{(T)}$ be the corresponding FPR gaps as defined in Eq. (\ref{eqn3.2.1}). This leaves us with a $2T$-dimensional Pareto frontier, which is hard to visualize and compare among multiple models. 

A straight-forward approach for measuring fairness-accuracy trade-off in multi-task learning is to project the $2T$-dimensional metric into a 2-dimensional (accuracy, fairness) metric by averaging accuracy metrics and fairness metrics across all tasks:
\begin{align}
\label{eqn4.1.2}
\overline{Err} = \frac{1}{T}\sum_{t=1}^T Err^{(t)}, 
\ \ \ \overline{FPRGap} = \frac{1}{T}\sum_{t=1}^T FPRGap^{(t)}.
\end{align}
However, naive averaging ignores the relative metric scale and baseline values for each task. We define a set of normalized average metric, \textbf{average relative fairness gap} ($ARFG$) and \textbf{average relative error} ($ARE$), which consolidates metric scales by comparing each multi-task metric against its single-task counterparts: 


\begin{dfn} 
\label{dfn4.1.1}
(Average relative fairness gap and average relative error). 
Let $Err_{S}^{(t)}$ and $FPRGap_{S}^{(t)}$ be the error rate and FPR gap for task $t$ using single-task learning with the same model architecture as in multi-task learning ($t=1,...,T$) and without any fairness remediation. We define Average Relative Fairness Gap (ARFG) and Average Relative Error (ARE) for a multi-task learning model as the average of relative FPR Gap and error rates against single-task baselines:
\begin{align}
\label{eqn4.1.3}
\begin{split}
ARFG &\coloneqq \frac{1}{T}\sum_{t=1}^T FPRGap^{(t)} / FPRGap_{S}^{(t)}, \\
ARE &\coloneqq \frac{1}{T}\sum_{t=1}^T Err^{(t)} / Err_{S}^{(t)}. 
\end{split}
\end{align}
\end{dfn}

Note that the single-task learning baselines $FPRGap_{S}^{(t)}$ and $Err_{S}^{(t)}$ are obtained using the same model architecture as in the multi-task learning case. Similar to the individual $FPRGap$ and error rates, \emph{low} values of $ARFG$ and $ARE$ indicate good overall fairness and accuracy. Table \ref{tab4.1.1} shows an example that compares these new definitions with naive averaging as in Eq. (\ref{eqn4.1.2}). We see that between the two hypothetical models with the same average error $\overline{Err}$, Model A has much lower $ARE$ than Model B. This is because $ARE$ considers relative changes rather than absolute values. Compared with single-task learning, Model A reduces Task 1 error by 5\% (from 0.40 to 0.38), Model B reduces Task 2 error by 50\% (from 0.04 to 0.02) while keeping the other task's error unchanged. When the two tasks are equally important, intuitively Model B is better than Model A as it achieves better relative error reduction. This is reflected in $ARE$ but not $\overline{Err}$. 

\begin{table}[h]
  \begin{center}
    \begin{tabular}{c|cccc}
      \hline\hline
                  & T1 Error & T2 Error & $\overline{Err}$ & $ARE$  \\
      \hline
      STL-T1          & 0.40       & -        & -      & -     \\
      STL-T2          & -          & 0.04     & -      & -     \\ 
      \hline
     MTL Model A      & 0.40       & 0.02     & \textbf{0.21}   & \textbf{0.75}     \\
     MTL Model B      & 0.38       & 0.04     & \textbf{0.21}   & \textbf{0.975}     \\
      \hline\hline 
    \end{tabular}
    \vspace{.1in}
    \caption{A hypothetical example comparing $ARE$ with $\overline{Err}$. \textbf{STL-T1/T2}: single-task learning for Task 1/Task 2, used to get the baseline error $Err_{S}^{(t)}$ for Task $t=1,2$; \textbf{MTL Model A/B}: Two hypothetical multi-task models for comparison.}
    \label{tab4.1.1} 
  \end{center}
  \vspace{-.3in}
\end{table}

Note that these average relative metrics $ARFG$ and $ARE$ are always positive, and can be either smaller or greater than 1 as multi-task learning could either improve or hurt accuracy or fairness of individual tasks as shown in Section \ref{section3.3}. $ARFG < 1$ ($ARE < 1$) suggests that multi-task learning reduces relative FPR gap (error) on average, and vice versa. The trade-off between $ARFG$ and $ARE$ depicts the Pareto frontier of accuracy and fairness while accounting for different metric scales for different tasks. We call it \emph{$ARFG$-$ARE$ Pareto frontier}.

\section{Method}

We first discuss the baseline method for adding fairness treatment to a multi-task learning model, which is a naive generalization of the single-task learning case. We focus on equal opportunity in this section, but the formulation and discussions can be generalized to other group fairness objectives such as equalized odds. 

\subsection{Baseline: Per-Task Fairness Treatment} 
\label{sec:baseline}

\textbf{Group Loss definitions}.
For single-task learning, equal opportunity (i.e.\ equalized FPR) entail matching the predictions over negative examples across groups, and can be accomplished through:
\begin{itemize}
    \item minimizing the correlation between group membership and the predictions over negative examples \cite{beutel2019fairness, beutel2019putting} (Eq.~\ref{eqn4.2.0a}); \item kernel-based distribution matching through Maximum Mean Discrepancy (MMD) over negative examples \cite{prost2019toward} (Eq.~\ref{eqn4.2.0b}); 
    \item minimizing FPR gap directly in the loss \cite{JMLR:v20:18-262, feldman2015certifying, Menon2018TheCO} (Eq.~\ref{eqn4.2.0c}). 
\end{itemize}

The way these methods work is intuitive: By closing the gap on the prediction differences on negative examples, the model is able to achieve similar FPR across all groups, thus optimizing towards equal opportunity. Therefore, we can define losses as essentially different measures of distance between the predictions on sensitive subgroups defined by $A$ over the \emph{subpopulation with negative examples} (i.e. $Y=0$), represented by $N$. To be precise, the \emph{population} version of these fairness losses are defined as:
\begin{subequations}
 \label{eqn4.2.0}
 \begin{align}
 & \F_{Corr}(f|N) \coloneqq Corr(f(x), A \,|\, Y=0), \label{eqn4.2.0a} \\
 & \F_{MMD}(f|N) \coloneqq MMD(\{{f(x)}|A=0\}, \{{f(x)}|A=1\} \,|\, Y=0), \label{eqn4.2.0b}\\
 & \F_{FPR}(f|N) \coloneqq |P(\hat{Y}=1|Y=0, A=0) - P(\hat{Y}=1|Y=0, A=1)|. 
 \label{eqn4.2.0c}
 \end{align}
\end{subequations}

\noindent \textbf{Empirical Loss definitions}.
Given the examples $\{(x_i, y_i, a_i)\}_{i=1}^n$ where $a_i$ is the value of the sensitive attribute $A$ on the $i$-th example, the \emph{empirical} equal opportunity fairness losses are computed as: 
\begin{subequations}
\label{eqn4.2.1}
\begin{align}
& \Fhat_{Corr}(f|N) = \hat{Corr}(\{f(x_i)\}, \{a_i\} \,|\, y_i=0), \label{eqn4.2.1a} \\
& \Fhat_{MMD}(f|N) = MMD(\{f(x_i): a_i=0\}, \{f(x_i): a_i=1\} \,|\, y_i=0), \label{eqn4.2.1b}\\
& \Fhat_{FPR}(f|N) = |\frac{\sum_i\mathbbm{1}[\hat{y_i}=1, y_i=0, a_i=0]}{\sum_i\mathbbm{1}[y_i=0, a_i=0]} - \frac{\sum_i\mathbbm{1}[\hat{y_i}=1, y_i=0, a_i=1]}{\sum_i\mathbbm{1}[y_i=0, a_i=1]}|,  \label{eqn4.2.1c}
\end{align}
\end{subequations}
where MMD distance \cite{gretton2008kernel} measures the distance between two groups of values based on kernel methods.

\noindent \textbf{Baseline Remediation}.
For remediation in single-task learning, the empirical loss function is a linear combination of the primary accuracy loss and the fairness loss:
\vspace{-0.05in}
\begin{equation}
  \label{eqn4.2.3}
  \Lhat_{STL}(f) = \Lhat(f) + \lambda \Fhat(f|N),
\end{equation}
where $\Lhat(f) = \frac{1}{n}\sum_{i=1}^n \L(f(x_i), y_i)$ is the empirical accuracy loss and $\Fhat(f|N)$ is the fairness loss of choice as defined in Eq. (\ref{eqn4.2.1}); The subscript STL stands for single-task learning, and $\lambda$ is a hyperparameter controlling the trade-off between accuracy and fairness. 

Note that while the primary loss $\Lhat(f)$ is computed for every example, the computation for fairness loss depends on the fairness notion of choice. For equal opportunity which minimizes FPR gap, only negative examples are included in the fairness loss computation (Eq. (\ref{eqn4.2.0}) and (\ref{eqn4.2.1})); For equalized odds which minimized both FPR gap and TPR gap, there will be two terms in the fairness loss $\Fhat(f|\cdot)$, with one computed on negative examples corresponding to minimizing FPR gap, and the other computed on positive examples corresponding to minimizing TPR gap.
 
A naive generalization of fairness treatments to multi-task learning is to apply the single-task fairness treatment to each task separately. For equal opportunity as an example, for every task we take its negative examples, compute the fairness loss, and add to the primary losses for the accuracy objectives in Eq. (\ref{eqn:3.1.2}). This yields a \emph{per-task} fairness treatment for multi-task learning: Using the notations in Section \ref{section3.1} where the task prediction functions are parameterized by $\theta = (\theta_{sh}, \theta_1,...,\theta_T)$ and dropping the dependency on $(x_i, y_i^t, a_i)$ for ease of notation, the total empirical loss equals:
\vspace{-0.05in}
\begin{equation}
  \label{eqn4.2.4}
  \vspace{-0.05in}
  \Lhat_{MTL}(\theta)=\sum_{t=1}^T w_t [\Lhat_t(\theta) + \lambda_t \Fhat_t(\theta)],
  \vspace{-0.05in}
\end{equation}
where $\Lhat_t(\theta)$ as in Eq. (\ref{eqn:3.1.2}) is the empirical loss for the accuracy objective (we call it ``accuracy los'' in the remainder of this paper) for task $t$, and $\Fhat_t(\theta)$ is the empirical loss for the fairness objective (we call it ``fairness loss'' in the remainder of this paper), and $\lambda_t$ is the fairness weight which controls the relative trade-off between accuracy and fairness for task $t$. 

\begin{figure}[tb]
        \centering
        \begin{subfigure}[b]{0.23\textwidth}
        \centering
        \includegraphics[width=\textwidth]{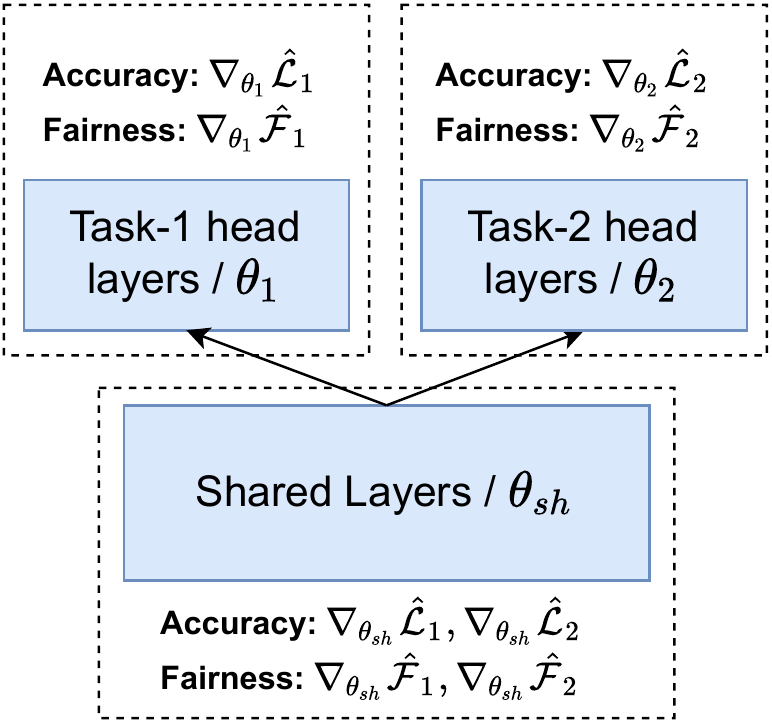}
        \caption{Baseline method.\label{fig4.2.1a}}
        \end{subfigure}
        \begin{subfigure}[b]{0.23\textwidth}
        \centering
        \includegraphics[width=\textwidth]{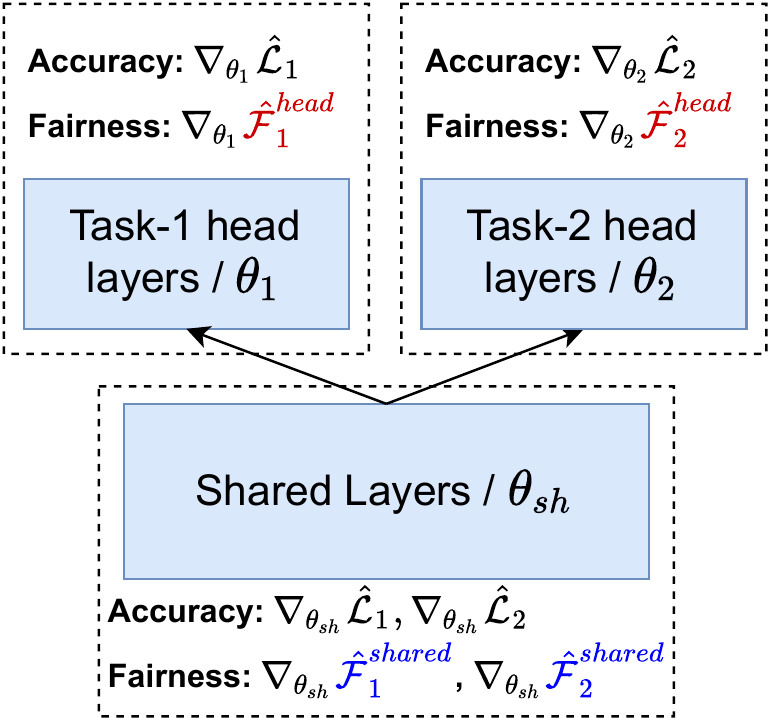}
        \caption{MTA-F method.\label{fig4.2.1b}}
        \end{subfigure}
        \vspace{-0.1in}
        \caption{Effective gradients on different model components.\label{fig4.2.1}}
\vspace{-0.1in}
\end{figure}

For a multi-task model as in Fig. \ref{fig:shared-bottom-mtl}, the shared layers (parameterized by $\theta_{sh}$) receive gradient updates from all task losses during training, while the task-specific head layers (parameterized by $\theta_1, ... , \theta_T$) effectively receive gradient updates only from its own task losses, as $\nabla_{\theta_{t}} \L(f(x;\theta_{sh}, \theta_{t'}), y) \equiv 0$ for any $t \neq t'$. Figure \ref{fig4.2.1a} illustrates this fact for the case of $T=2$ tasks. 


\subsection{Our Method: Multi-Task-Aware Fairness Treatment (MTA-F)}
\label{sec:ourmethod}
We now present our multi-task-aware fairness treatment framework for improving fairness-accuracy trade-off for multi-task learning. We start with equal opportunity as the fairness objective and generalize to equalized odds later.

\begin{figure}[tb]
\begin{subfigure}{.48\textwidth}
  \centering
  \includegraphics[width=.8\linewidth]{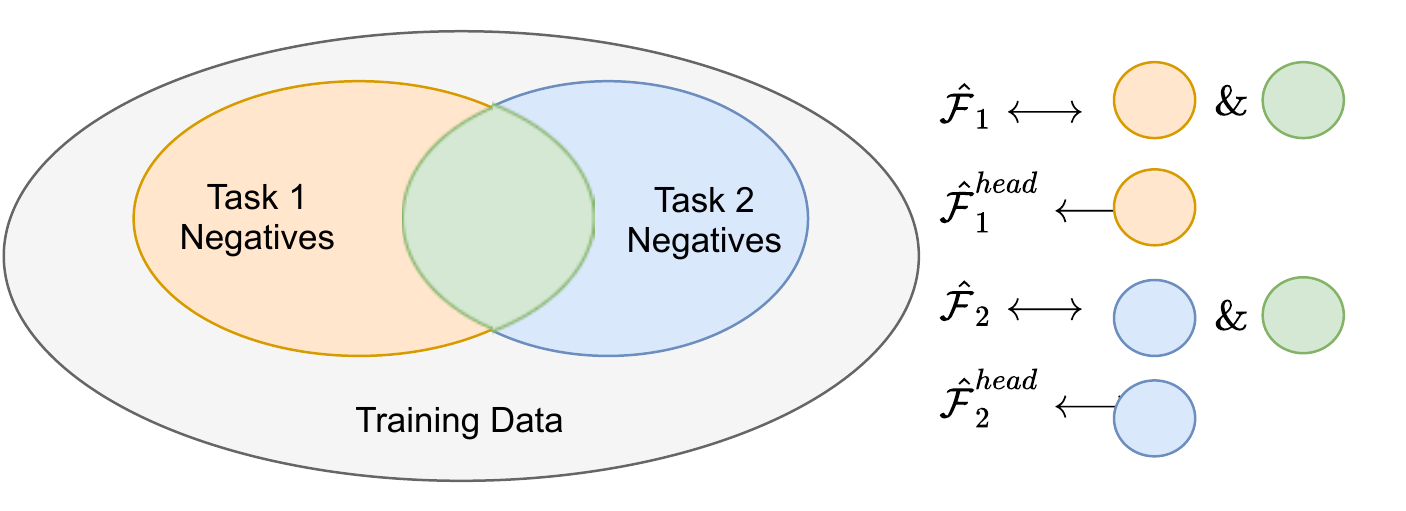}  
  \caption{Fairness loss decomposition with MTA-F.}
  \label{fig4.3.1a}
\end{subfigure}
\newline 
\vspace{2mm}
\begin{subfigure}{.48\textwidth}
  \centering
  \includegraphics[width=1.0\linewidth]{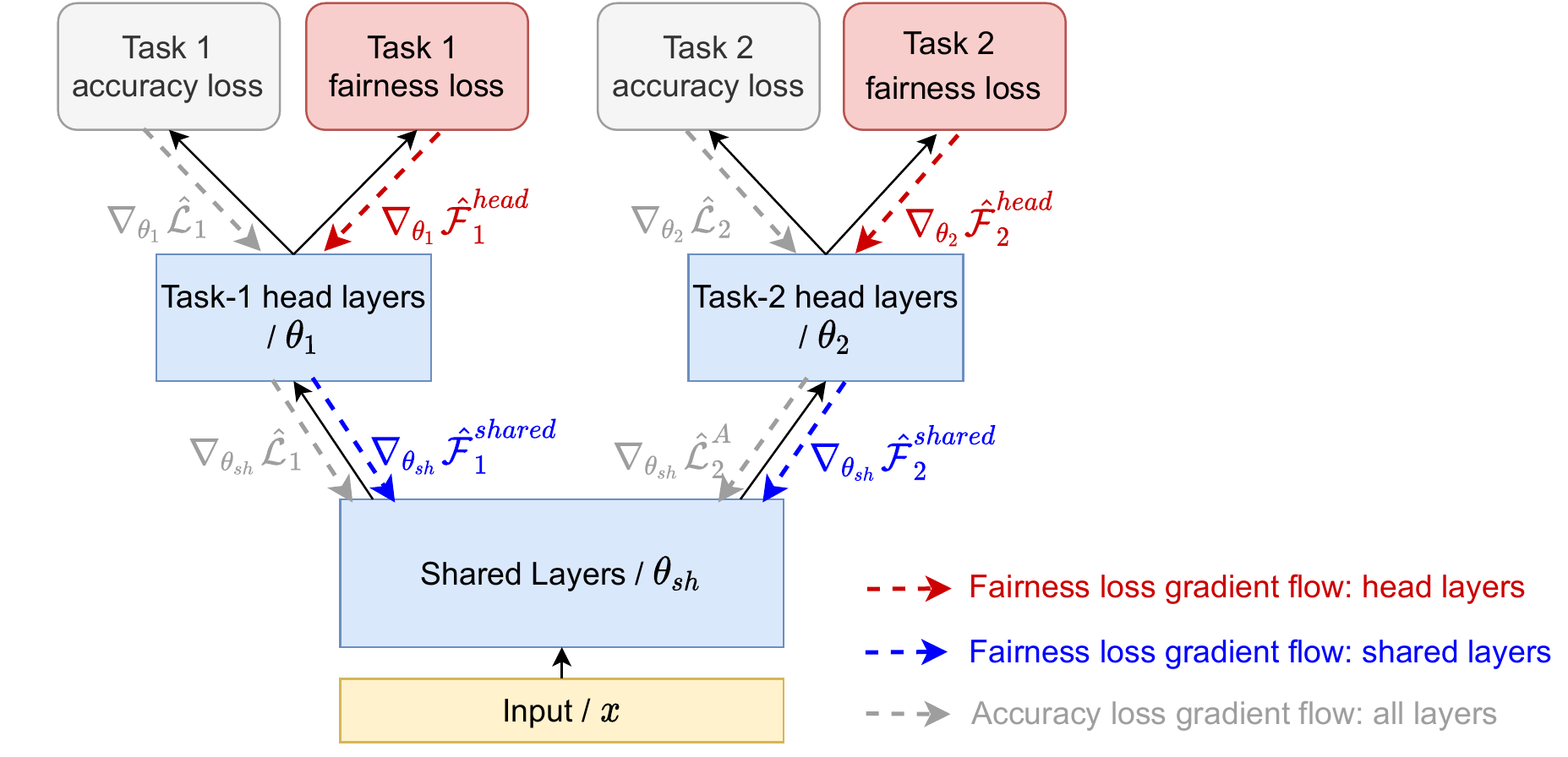}  
  \caption{Backpropagation with MTA-F: We backpropagate task-specific fairness losses $\Fhat_t^{head}$ to head layers, and the remaining fairness loss $\Fhat_t^{shared}$ to shared layers ($t=1,2$).}
  \label{fig4.3.1b}
\end{subfigure}
\vspace{-0.1in}
\caption{Multi-task-aware fairness treatment (MTA-F).}
\label{fig4.3.1}
\vspace{-0.1in}
\end{figure}

The fairness loss is computed on the negative examples of each task in order to minimize FPR gap towards equal opportunity. Let $N_1, ..., N_T$ be the sets of negative examples for each task, which may or may not overlap. In the example with 2 tasks, the fairness loss acting on the shared bottom layer is computed on the union of negative examples for either Task 1 or Task 2 ($N_1\cup N_2$). However, the examples that are negative on Task 1 but positive on Task 2 ($N_1\cap N_2^c$, the orange region in Fig. \ref{fig4.3.1a}) should only be relevant to the fairness goal of Task 1; likewise the examples that are negative on Task 2 but positive on Task 1 ($N_2\cap N_1^c$, the blue region in Fig. \ref{fig4.3.1a}) should only be relevant to the fairness goal of Task 2. In the baseline method in Section \ref{sec:baseline}, the fairness loss does not distinguish those examples and include all of them in the fairness loss computation for both shared layers and head layers, which intuitively is a suboptimal use of the model capacity. 
Inspired by this observation, we propose a decomposition and redistribution of fairness treatments for multi-task learning: For every task $t$, obtain its fairness loss $\Fhat_t$ computed on the negatives $N_t$ as in the baseline method. Then compute the fairness loss on the exclusive negatives $N_t \cap (\cap_{k \neq t} N_k^c)$, i.e. the examples that are negative only on task $t$ but not the rest, and apply them to the head layers; And leave the remaining fairness loss to the shared layers.

Formally, define $N_t \coloneqq \{(x_i, y_i^1, ... , y_i^T, a_i): y_i^t = 0\}$ as subsets of examples that have negative labels on task $t$ for $ t=1,...,T $. Decompose the total fairness loss $\Fhat(\theta | N_t)$ computed on $N_t$ as task-specific fairness loss $\Fhat_t^{head}$ and shared fairness loss $\Fhat_t^{shared}$: 

\begin{equation}
\label{eqn4.3.1}
\begin{aligned}
\Fhat_t^{head}(\theta) &\coloneqq \Fhat(\theta \,|\,  N_t \cap (\mathop{\cap}_{k \neq t} N_k^c)), \\
\Fhat_t^{shared}(\theta) &\coloneqq \Fhat(\theta | N_t) - \Fhat_t^{head}(\theta),
\end{aligned}
\end{equation}
for every task $t=1,...,T$, where $\Fhat_t$ is any fairness loss function as defined in Eq. (\ref{eqn4.2.1}). During training, we backpropagate task-specific fairness losses $\Fhat_t^{head}$ to the head layers ($\theta_t$) and the remaining fairness losses $\Fhat_t^{shared}$ to the shared layers ($\theta_{sh}$). 

Figure \ref{fig4.3.1b} illustrates the proposed idea on a 2-task model and how the gradient flows during backpropagation. Note that $\Fhat_t^{shared}$ is not equivalent to the fairness loss computed on the shared negative examples (green region in Fig \ref{fig4.3.1a}) as most fairness losses are not additive. The number of fairness losses with this decomposition scales linearly with $T$. Figure \ref{fig4.2.1} illustrates the difference in fairness gradients between the proposed method and the baseline method. Because it is considering the labels of other tasks when designing the fairness loss for each task, we call this method \textbf{multi-task-aware fairness treatment (MTA-F)}. 



The multi-task learning update step described in Algorithm \ref{algo4.3.1} is applicable to any problem that uses gradient-based optimization. For task $t$, in addition to the accuracy loss weight $w_t$ and fairness loss weight $\lambda_t$ as in the baseline algorithm in Section \ref{sec:baseline}, we introduce an additional hyperparameter $r_t > 0$ which controls the relative ratio of the fairness loss weight between the head layers and the shared layers. 

\setlength{\textfloatsep}{10pt}
\begin{algorithm}[ht]
\SetAlgoNoLine
\DontPrintSemicolon

\KwIn{Mini-batch ${\{(x_i, y^1_i, ... ,y^T_i)\}}_{i=1}^n$, model parameters $\theta = (\theta_{sh}, \theta_1,...,\theta_T)$, task weights $\{w_t\}_{t=1}^T$, fairness weights $\{\lambda_t\}_{t=1}^T$, head-to-shared ratio $\{r_t\}_{t=1}^T$, and learning rate $\eta$} 
 \For{$t=1,...,T$}{
 $\Lhat_t(\theta_{sh}, \theta_t) = \frac{1}{n}\sum_{i=1}^n \L_t(f_t(x_i;\theta_{sh}, \theta_t), y^t_i)$ \; \Comment*[r]{Compute accuracy losses} 
 $\Fhat_{t}^{head}(\theta_{sh}, \theta_t)$ and $\Fhat_{t}^{shared}(\theta_{sh}, \theta_t)$ as in Eq. (\ref{eqn4.3.1}) / (\ref{eqn4.3.2}) \; \Comment*[r]{Compute fairness losses} 
 $\theta_t = \theta_t - \eta w_t [\nabla_{\theta_t} \Lhat_t(\theta_{sh}, \theta_t) + \lambda_t r_t \nabla_{\theta_t} \Fhat_{t}^{head}(\theta_{sh}, \theta_t)]$\;  \Comment*[r]{Gradient descent on head parameters} 
 } 
$\theta_{sh} = \theta_{sh} - \eta  \{ \sum_{t=1}^T w_t [\nabla_{\theta_{sh}} \Lhat_t(\theta_{sh}, \theta_t) $ \;
 \nonl\qquad\qquad\qquad\qquad\qquad\qquad $ + \; \lambda_t \nabla_{\theta_{sh}} \Fhat_t^{shared}(\theta_{sh}, \theta_t)]  \}$\; \Comment*[r]{Gradient descent on shared parameters}
\KwOut{Updated model parameters $\theta = (\theta_{sh}, \theta_1,...,\theta_T)$}
\caption{MTA-F Update Rule}
\label{algo4.3.1}
\end{algorithm}

Section \ref{experiments} shows the effectiveness of MTA-F in achieving a better fairness-accuracy trade-off on real-world datasets. The fact that a simple decomposition and redistribution of the fairness losses improves the Pareto efficiency in multi-task learning is not surprising to us. MTA-F enables different parts of the model to address the fairness concerns across multiple tasks in a targeted way. Head layers address fairness issues that are \emph{specific} to the task itself, while shared layers address fairness issues that are \emph{common} to multiple tasks. The mediation leads to a more efficient allocation of model capacity for fairness objectives, which effectively ``saves'' the model more capacity for accuracy objectives, thus enabling a better fairness-accuracy trade-off. On a related note, existing studies show that multi-task models of the same capacity can have different generalization performance for accuracy objectives, by implicitly manipulating the information flow in different layers \cite{meyerson2018pseudo,wang2020small}.

We point out that although we describe the formulation of MTA-F for optimizing equal opportunity, it can be easily generalized to other fairness metrics such as TPR gap for equal opportunity (instead of FPR gap), or both FPR gap and TPR gap for equalized odds. When TPR gap is the fairness goal, the fairness losses are computed on the positive examples (instead of negative examples) of each task. In the case for equalized odds where both FPR gap and TPR gap are of interest, we just need to add two addition fairness loss terms in $\Fhat_t^{head}$ and $\Fhat_t^{share}$ in Eq. (\ref{eqn4.3.1}) as:
\begin{equation}
\label{eqn4.3.2}
\begin{aligned}
\Fhat_t^{head}(\theta) &= \Fhat(\theta \,|\,  N_t \cap (\mathop{\cap}_{k \neq t} N_k^c)) + \Fhat(\theta \,|\,  P_t \cap (\mathop{\cap}_{k \neq t} P_k^c)), \\
\Fhat_t^{shared}(\theta) &= \Fhat(\theta | N_t) + \Fhat(\theta | P_t) - \Fhat_t^{head}(\theta),
\end{aligned}
\end{equation}
where $P_t \coloneqq \{(x_i, y_i^1, ... , y_i^T, a_i): y_i^t = 1\}$ is the set for \emph{positive} examples of task t ($1\leq t\leq T$), and $\Fhat$ is the fairness loss of choice. Everything else works exactly the same as in Algorithm \ref{algo4.3.1}. Therefore MTA-F is a general framework for improving Pareto efficiency for multi-task learning toward fairness goals including equal opportunity and equalized odds, and works with different fairness loss functions. MTA-F also works when the sensitive attribute is available only for a fraction of the training data, for which the fairness loss computation and decomposition in Eq. (\ref{eqn4.3.1}) is done on the subset of training data where the sensitive attribution information is available.



\section{Experiments}
\label{experiments}

In this section, we present experiment results of MTA-F on three real-world datasets to demonstrate its effect on improving fairness-accuracy trade-off for multi-task learning problems. We first introduce the datasets and the experiment setup, then we present the results on the Pareto frontier of ARFG and ARE metrics defined in Section \ref{sec:measuring}, along with the fairness and accuracy metrics per task.

\subsection{Experiment Setup}

\textbf{Datasets:} We test on three widely used datasets with known fairness concerns: the UCI-Adult dataset, the German Credit Data\footnote{https://archive.ics.uci.edu/ml/datasets/statlog+(german+credit+data)} and the LSAC Law School dataset\cite{wightman1998lsac}. We formulate the multi-task learning problems by picking two binary tasks for each of them: $Income >\$50,000$ and $Capital\; Gain > 0$ for UCI-Adult dataset, \emph{Good loans} and \emph{High credit} for the German Credit Data, and \emph{Pass bar exam} and \emph{High first year GPA} for the LSAC Law School dataset. Details on the datasets and task constructions can be found in Appendix \ref{sup:datasets}. Gender is treated as the sensitive attribute for all three datasets.


\noindent \textbf{Treatments:} The methods we compare in the experiments are: 
\begin{itemize}
    \item \textbf{Vanilla MTL}: The naive multi-task learning without any fairness remediation, i.e. only minimizing combination of accuracy losses as in Eq. (\ref{eqn:3.1.2}); 
    \item \textbf{Baseline}: The baseline  treatment described in Section \ref{sec:baseline}, with the minimization objective as in Eq. (\ref{eqn4.2.4}); 
    \item \textbf{MTA-F}: Our proposed multi-task-aware fairness treatment described in Section \ref{sec:ourmethod} and Algorithm \ref{algo4.3.1}. 
\end{itemize}

\noindent \textbf{Fairness Loss:} In order to test our method on different fairness loss functions, we adopt MMD loss $\F_{MMD}$, FPR Gap loss $\F_{FPR}$ and correlation loss $\F_{Corr}$ as defined in Eq. (\ref{eqn4.2.0}) as the fairness loss on the three datasets respectively. The losses are computed on negative examples, optimizing toward equal opportunity as the fairness goal which is measured by FPR gap. Cross-entropy loss is used for all accuracy losses. 

\noindent \textbf{Architecture and tuning:} For all three datasets, we adopt the standard multi-task architecture with shared representation learning and task sub-networks as in Fig. \ref{fig:shared-bottom-mtl}. For each method, the same set of hyperparameter values is used across different fairness and accuracy weights. Details on the model architecture, tuning and the code can be found in Appendix \ref{sup:experiment_setup}. Once the hyperparameters are selected, we perform 20k runs for each method with varying task fairness weights and loss weights, evaluate each of them on test dataset, and report the Pareto frontier of the test metrics.

\subsection{Results}
\label{sec:results}

In Table \ref{ar_metrics}, we first present the numerical results by reporting one point on the $ARFG$-$ARE$ Pareto frontier and read the corresponding $ARFG$ and $ARE$ metrics. For each method, the point is chosen to be such that its $ARE$ value is closest to the midpoint of the interval of $ARE$ values from all runs of that method. The corresponding per-task fairness and accuracy metrics are reported in Table \ref{per_task_metrics}. 

\begin{table}[hbt]
  \begin{center}\small
    \begin{tabular}{c|cc|cc|cc}
      \hline\hline
       Dataset   &\multicolumn{2}{c|}{UCI-Adult} & \multicolumn{2}{c|}{German Credit} & \multicolumn{2}{c}{LSAC Law School} \\
      \hline
        Metric          & $ARFG$ & $ARE$ & $ARFG$ & $ARE$ & $ARFG$ & $ARE$  \\
      \hline
     Vanilla MTL     & 0.3444  &  1.1040     & 0.1336  &  0.8367  & 0.3497  &  0.9778      \\
     
     Baseline      & 0.0871       & 1.1032  & 0.0999     & 0.8356 & 0.1126       & 0.9864        \\

     \textbf{MTA-F}       
     & \textbf{0.0437}     & \textbf{1.0820}
      & \textbf{0.0364}       & \textbf{0.8264}   
     & \textbf{0.0310}     & \textbf{0.9731}         \\

      \hline\hline
    \end{tabular}
    \vspace{1mm}
    \caption{Average relative fairness gap ($ARFG$) and average relative error ($ARE$) on \textbf{UCI-Adult}, \textbf{German Credit Data} and \textbf{LSAC Law School} datasets, as defined in Section \ref{sec:measuring}. Lower metric values indicate better overall fairness / accuracy across all tasks.}
    \label{ar_metrics}
  \end{center}
  \vspace{-0.4in}
\end{table}

\begin{table}[hbt]
  \begin{center}\small
    \begin{tabular}{p{0.8cm}l|cccc}
      \hline\hline
     &  & $T_1$ $Err$ & $T_1$ $FPRGap$  & $T_2$ $Err$ & $T_2$ $FPRGap$ \\
    \hline
   \multirow{3}{0.8cm}{UCI-Adult} & Vanilla MTL     & 0.1911  & 0.0715  & 0.1359 & 0.0091   \\

    & Baseline     & 0.1938 & 0.0186 & 0.1336  & 0.0020    \\

    & \textbf{MTA-F}    & \textbf{0.1891} & \textbf{0.0083} & \textbf{0.1319}  & \textbf{0.0016}    \\
      \hline
    \multirow{3}{0.8cm}{German Credit} & Vanilla MTL     & 0.205  & 0.0150  & 0.220 & 0.0084   \\

    & Baseline     & 0.255 & 0.0879 & \textbf{0.180}  & 0.0069    \\

   &  \textbf{MTA-F}    & \textbf{0.200} & \textbf{0.0033} & 0.220  & \textbf{0.0034}    \\
     
      \hline
    \multirow{3}{0.8cm}{LSAC Law School} & Vanilla MTL     & 0.1555  & 0.0503  & \textbf{0.1565} & \textbf{0.0004}   \\

    & Baseline     & 0.1568 & 0.0119 & 0.1580  & 0.0006    \\

    & \textbf{MTA-F}    & \textbf{0.1540} & \textbf{0.0015} & \textbf{0.1565}  & \textbf{0.0004}    \\  
      \hline\hline
      
    \end{tabular}
  \end{center}
 \caption{Per-task metrics for \textbf{UCI-Adult, German Credit Data} and \textbf{LSAC Law School} datasets.}
 \label{per_task_metrics}
 \vspace{-0.2in}
\end{table}

\subsubsection{UCI-Adult Results}
\label{section5.1}
Here we adopt the MMD fairness loss as defined in Eq. (\ref{eqn4.2.0b}) and (\ref{eqn4.2.1b}) with Gaussian kernel.

\begin{figure}[tb]
        \centering
        \begin{subfigure}[b]{0.23\textwidth}
                \centering
                \includegraphics[width=\textwidth]{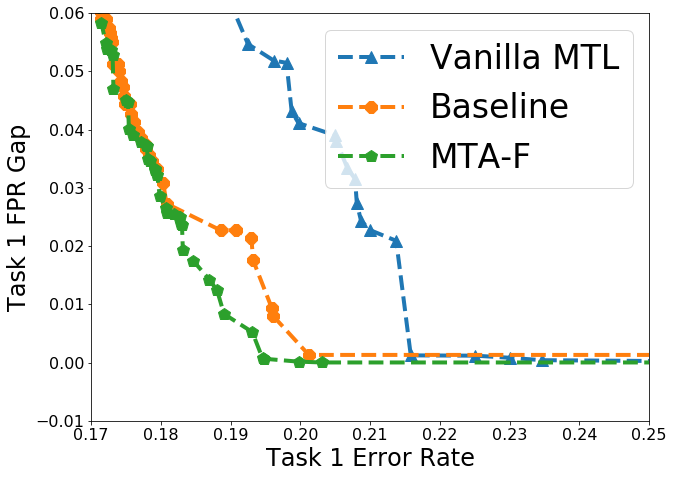}
                \caption{Fairness-accuracy Pareto frontier for \textbf{Task 1}.\label{fig5.1.1a}}
        \end{subfigure}\hfill
        \begin{subfigure}[b]{0.23\textwidth}
                \centering
                \includegraphics[width=\textwidth]{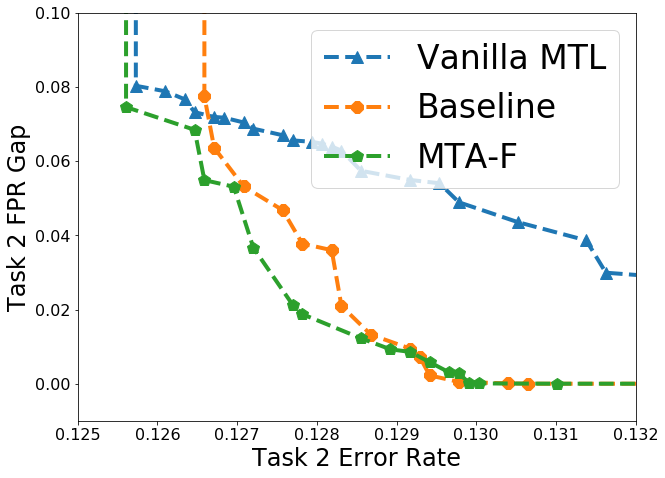}
                \caption{Fairness-accuracy Pareto frontier for \textbf{Task 2}.\label{fig5.1.1b}}
        \end{subfigure}
        \begin{subfigure}[b]{0.23\textwidth}
                \centering
                \includegraphics[width=\textwidth]{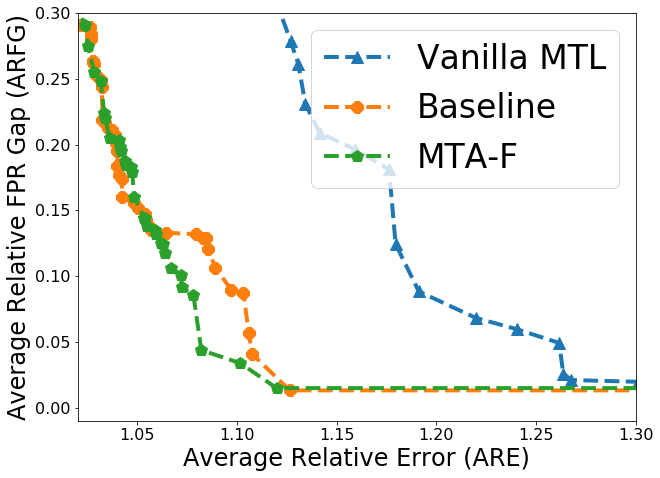}
                \caption{$ARFG$-$ARE$ Pareto frontier.\label{fig5.1.1c}}
        \end{subfigure}
        \vspace{-0.1in}
        \caption{Pareto frontiers on UCI-Adult dataset. Lower-left indicates better Pareto optimality, i.e. better fairness-accuracy trade-off. \label{fig5.1.1}}
\vspace{-0.2in}
\end{figure}

Figure \ref{fig5.1.1a}-\ref{fig5.1.1b} show the fairness-accuracy Pareto frontiers of each task separately, and that MTA-F is able to improve the per-task fairness-accuracy trade-off for both tasks. Note that these Pareto frontiers are only \emph{marginal} frontiers in the sense that model runs that are Pareto optimal for Task 1 (i.e. points on Fig. \ref{fig5.1.1a}) may not correspond to Pareto optimal runs for Task 2 (i.e. points on Fig. \ref{fig5.1.1b}). Therefore, better per-task fairness-accuracy trade-offs do not guarantee overall better fairness-accuracy trade-off across all tasks. 

On the contrary, the $ARFG$-$ARE$ Pareto frontier proposed in Section \ref{sec:measuring}, which consists of Pareto optimal runs for $ARFG$ and $ARE$, is able to capture the trade-off between overall fairness and overall accuracy for multi-task learning problems. Figure \ref{fig5.1.1c} shows the $ARFG$-$ARE$ Pareto frontier. We see that MTA-F is also able to improve the overall fairness-accuracy trade-off across all tasks, compared with Baseline fairness treatment and Vanilla MTL. 

As we can see from Table \ref{ar_metrics}, both Baseline and MTA-F are able to significantly reduce the $ARFG$ metric compared with Vanilla MTL, confirming the effectiveness of fairness treatments using MMD loss. MTA-F achieves the lowest $ARFG$ value. Note that the $ARE$ metrics for all methods are slightly greater than 1, meaning that training both tasks together via multi-task learning generate slightly worse accuracy metrics than single-task learning, which suggests potential task training conflicts due to limited model capacities. However, MTA-F \emph{Pareto dominates} Vanilla MTL and Baseline in that it has lower values in both $ARFG$ and $ARE$ metrics, indicating a better overall fairness-accuracy trade-off across all tasks.

\subsubsection{German Credit Data Results}
\label{section5.2}
\begin{figure}[tb]
        \centering
        \begin{subfigure}[b]{0.23\textwidth}
                \centering
                \includegraphics[width=\textwidth]{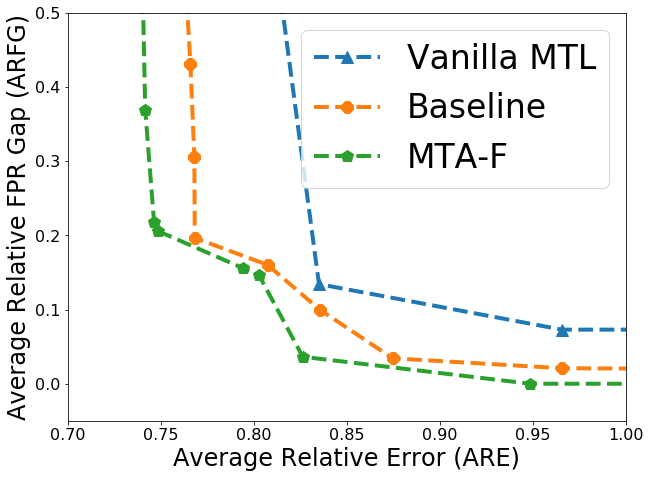}
                \caption{German Credit Data.\label{arfg-are-pareto-frontier-a}}
        \end{subfigure}
        \hfill
        \begin{subfigure}[b]{0.23\textwidth}
                \centering
                \includegraphics[width=\textwidth]{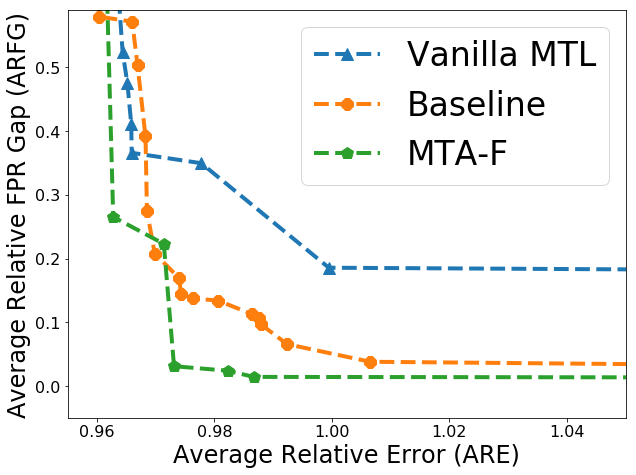}
                \caption{LSAC Law School.\label{arfg-are-pareto-frontier-b}}
        \end{subfigure}
        \vspace{-0.1in}
        \caption{$ARFG$-$ARE$ Pareto frontiers. \label{arfg-are-pareto-frontier-}}
        \vspace{-0.1in}
\end{figure}

We adopt FPR gap as the fairness loss $\F_{FPR}$ as defined in Eq. (\ref{eqn4.2.0c}) for equal opportunity on German Credit Data. Figure \ref{arfg-are-pareto-frontier-a} and Table \ref{ar_metrics} and \ref{per_task_metrics} show the $ARFG$-$ARE$ Pareto frontier and numerical values, respectively, where the numerical results correspond to one of the points on the Pareto frontier chosen in the same way as described in the beginning of Section \ref{sec:results}. 
We see that different from the results on UCI-Adult dataset, the $ARE$ metrics are smaller than 1 on German Credit Data for all three methods, which means that the two tasks benefit from each other when trained together on a share model architecture. \citet{wu2020understanding} call this \emph{positive transfer}. The results show that with a different fairness loss function, MTA-F is still able to improve both $ARFG$ and $ARE$ over Baseline and Vanilla MTL, therefore improving the overall fairness-accuracy trade-off.

\subsubsection{LSAC Law School Dataset Results}
\label{section5.3}
Here we test another choice for fairness loss: the correlation loss $\F_{Corr}$ defined in Eq. (\ref{eqn4.2.0a}). Figure \ref{arfg-are-pareto-frontier-b} shows the $ARFG$-$ARE$ Pareto frontier and Table \ref{ar_metrics} and \ref{per_task_metrics} show the numerical results. On LSAC Law School dataset, training both tasks together with multi-task learning achieves on-par accuracy results as the single-task learning case ($ARE$ metrics are close to 1). Our MTA-F method is able to reduce $ARFG$ significantly while keeping ARE relatively the same, suggesting improved Pareto fairness compared with other methods.

\section{Conclusion}
\label{conclusion}

In this paper, we present insights and improvements for the understudied trade-off between fairness and accuracy in multi-task learning. Notably, the accuracy goals across multiple tasks, which current multi-task learning algorithms optimize for, may not align well with the fairness goals. To understand and measure the overall fairness-accuracy trade-off as well as the trade-offs across tasks in a multi-task learning setting, we propose a new set of metrics, average relative fairness gap ($ARFG$) and average relative error ($ARE$), which quantifies the multi-dimensional trade-off in multi-task settings where fairness is a concern. We then propose a data-dependent \emph{multi-task-aware fairness treatment (MTA-F)}, which adaptively decouples the fairness treatments in standard multi-task model architectures based on inter-task label distributions. MTA-F works with a number of fairness loss functions toward equal opportunity and equalized odds. Experimental results on three real-world datasets demonstrate the effectiveness of MTA-F in improving the fairness-accuracy trade-off for multi-task applications.

\bibliographystyle{ACM-Reference-Format}
\bibliography{reference}

\newpage
\appendix
\section{Appendix}


\subsection{Datasets}
\label{sup:datasets} 

\noindent \textbf{CelebA Dataset:} The CelebA dataset\footnote{http://mmlab.ie.cuhk.edu.hk/projects/CelebA.html} \cite{liu2015deep} contains public domain images of public figures. It has about 200,000 $64\times64$ face images of celebrities, each with 5 landmark locations, and 40 binary attributes including gender. We pick two out of the 40 attributes, \emph{Attractive} and \emph{Smiling} as the two tasks for the multi-task problem. The dataset is randomly split into training data of size 162,770 and test data of size 19,867.

\noindent \textbf{UCI-Adult Dataset:} 
The UCI-Adult dataset\footnote{https://archive.ics.uci.edu/ml/datasets/adult} contains census information with the training / test data containing information of 32,561 / 16,281 individuals. We pick the default binary label of the dataset, $Income >\$50,000$ or $\leq\$50,000$ as Task 1; and another quantity $Capital\; Gain$ from the dataset, transform it into a binary label $Capital\; Gain > 0$ or $= 0$ as Task 2, to formulate a multi-task problem. The rest of the attributes, including age, working hours per week, education, occupation and race, are used as input features to the model. We use the original train/test split with the dataset.

\noindent \textbf{German Credit Data:} 
The German Credit Data\footnote{https://archive.ics.uci.edu/ml/datasets/statlog+(german+credit+data)} provides a set of attributes for 1000 individuals, including credit history, credit amount, gender, age, etc., and the corresponding credit risk for that individual. We construct two binary tasks from the dataset: \emph{Good loans vs. Bad loans} as Task 1, and $Credit\; amount > 2000$ or $\leq 2000$ as Task 2. There are 16 attributes that we use as features. We randomly split the dataset into training data of size 800 and test data of size 200.

\noindent \textbf{LSAC Law School:} 
The LSAC dataset\footnote{http://www.seaphe.org/databases.php} \cite{wightman1998lsac} is generated from the survey conducted by the Law School Admission Council in the United States. It contains information on 21,790 law students such as their entrance exam scores (LSAT), their undergrad grade-point average (GPA) collected prior to law school, gender and race etc. For each student, we predict whether they passed the bar exam as Task 1, and whether they have a high first year average grade (defined by z-score > 0) as Task 2. The dataset is randomly split into training data of size 18585 and test data of size 7966.


\subsection{Experiment Setup}
\label{sup:experiment_setup} 

\subsubsection{Architecture}
\label{sup:architecture} 
For all datasets used in the paper, we adopt the standard multi-task architecture with shared representation learning and task sub-networks as in Fig. \ref{fig:shared-bottom-mtl}. For the CelebA dataset in Section \ref{section3.3}, we use ResNet-18 \cite{he2016deep} without the final layers as a shared representation, and two fully connected layers of size [1000, 500] as head layers for each task. For UCI-Adult dataset, the model has a single shared hidden layer with 64 shared hidden units and task-specific fully-connected hidden layers of size 32 with ReLU activation for task-specific networks, where categorical features are encoded as 40-dimensional embeddings. For German Credit Data and LSAC dataset, we pick a smaller architecture due to their smaller sample sizes: a single shared hidden layer of size 32 and single task-specific hidden layer of size 16 for each task, both with ReLU activations. Categorical features are encoded as 10-dimensional embeddings.

\subsubsection{Tuning}
\label{sup:tunin} 
The same set of hyperparameter values is used for each method, across different fairness and accuracy weights. 
Adagrad optimizer is used with learning rate $lr$ and number of epochs as hyperparameters. 

We use batch size of 512 for CelebA, UCI-Adult and LSAC dataset, and 128 for German Credit Data. 
For MTA-F, there is one more set of hyperparameters, $\{r_1, r_2\}$ which controls the ratio between task-specific fairness losses $\F_t^{head}$ and shared fairness losses $\F_t^{share}$ for Task $t=1,2$. 
To ensure we are not over-tuning MTA-F over other methods due to its additional hyperparameters, for all methods, hyperparameters (i.e. $\{lr, epochs\}$ for Vanilla MTL and Baseline, and $\{lr, epoch, r_1, r_2\}$ for MTA-F) are tuned over the \emph{same} number of runs (50k). 
During the 50k tuning runs, we vary the Task 1 fairness loss weights $\lambda_1$, $\lambda_2$ in $[0,5]$ and accuracy loss weights $w_1$ and $w_2=1-w_1$ in $[0,1]$. 

Once the hyperparameters are selected, we perform another 20k runs for each method with task fairness weights and loss weights varying as above, evaluate each of them on test dataset, and report the Pareto frontier of the test metrics (i.e. the Pareto optimal solutions from the 20k runs) including per-task fairness (FPR gap) and accuracy (classification error rate), as well as the ARFG and ARE metrics defined in Section \ref{sec:measuring}.

\subsubsection{Code}
\label{sup:links}
The code for the experiments in Section \ref{experiments} is available at: \url{https://github.com/kdd-2021-yuyan/pareto-fairness-mtl}.  

\end{document}